\def\year{2023}\relax
\documentclass[letterpaper]{article} 
\usepackage{aaai}  

\usepackage[hyphens]{url}  
\usepackage{graphicx} 
\usepackage{graphicx}  

\usepackage{enumerate}   
\usepackage{hyperref}

\usepackage{xcolor}

\usepackage{natbib}

\usepackage{times}
\usepackage{helvet}
\usepackage{courier}

\usepackage{mathtools}
\usepackage{amsfonts}

\usepackage{etoolbox}
\usepackage[linesnumbered,vlined,ruled,commentsnumbered]{algorithm2e}
\SetKwRepeat{Do}{do}{while}

\SetVlineSkip{0.2em}

\SetInd{0.2em}{0.7em}

\DontPrintSemicolon{}

\SetKwIF{If}{ElseIf}{Else}{if}{}{else if}{else}{endif}

\SetAlgoSkip{}

\SetAlCapHSkip{0cm}

\SetAlgoCaptionLayout{small}
\SetAlFnt{\small}

\SetArgSty{textnormal}

\SetKwComment{Comment}{$\triangleright$\ }{}

\makeatletter
\patchcmd\algocf@Vline{\vrule}{\vrule \kern-0.4pt}{}{}
\patchcmd\algocf@Vsline{\vrule}{\vrule \kern-0.4pt}{}{}
\makeatother

\usepackage{subcaption}
\usepackage[capitalise]{cleveref}

\DeclareRobustCommand{\abbrevcrefs}{%
\crefname{algorithm}{Alg.}{Algs.}%
}

\DeclareRobustCommand{\cshref}[1]{{\abbrevcrefs\cref{#1}}}

\usepackage{blindtext}

\usepackage{tikz}

\usepackage{tikzscale}

\usepackage{pgfplots}

\usepackage{pgfplotstable}

\usetikzlibrary{calc}

\usetikzlibrary{positioning}

\usepgfplotslibrary{statistics}

\usepgfplotslibrary{fillbetween}

\pgfplotsset{
  compat=1.15,
  mps basic/.style={
    xlabel near ticks,
    xlabel style={font=\footnotesize},
    ylabel near ticks,
    ylabel style={font=\tiny},
    xmajorgrids,
    major x grid style={dotted},
    ymajorgrids,
    major y grid style={dotted},
    tick label style={font=\tiny}
  },
  mps scientific x/.style={
    x tick label style={
      /pgf/number format/sci,
      font=\tiny
    }
  },
  mps scientific y/.style={
    y tick label style={
      /pgf/number format/sci,
      font=\tiny
    }
  },
  mps fixed x/.style={
    x tick label style={
      /pgf/number format/.cd,
      fixed,
      fixed zerofill,
      precision=6,
      /tikz/.cd,
      font=\tiny
    }
  },
  mps fixed y/.style={
    y tick label style={
      /pgf/number format/.cd,
      fixed,
      fixed zerofill,
      precision=6,
      /tikz/.cd,
      font=\tiny
    }
  },
  /pgfplots/myylabel absolute/.style={%
      /pgfplots/every axis y label/.style={at={(0,0.5)},xshift=#1,rotate=90,align=center},
      /pgfplots/every y tick scale label/.style={
        at={(0,1)},above right,inner sep=0pt,yshift=0.3em
      }
   }
}

\usepackage{xcolor}
\definecolor{espblack}{RGB}{0,0,0}
\definecolor{espwhite}{RGB}{255,255,255}
\definecolor{espgray}{RGB}{206,206,206}
\definecolor{esplightgray}{RGB}{224,224,224}
\definecolor{espdarkgray}{RGB}{168,168,168}
\definecolor{espsomewhatdarkgray}{RGB}{130,130,130}
\definecolor{espverydarkgray}{RGB}{100,100,100}
\definecolor{espblue}{RGB}{11,93,174}
\definecolor{esplightblue}{RGB}{59,175,236}
\definecolor{espdarkblue}{RGB}{6,26,64}
\definecolor{espred}{RGB}{206,62,21}
\definecolor{esplightred}{RGB}{206,62,21}
\definecolor{espdarkred}{RGB}{61,19,8}
\definecolor{espyellow}{RGB}{232,163,26}
\definecolor{espgreen}{RGB}{100,161,27}
\definecolor{esplightgreen}{RGB}{149,198,35}
\definecolor{espdarkgreen}{RGB}{49,92,43}
\definecolor{esppurple}{RGB}{106,20,125}
\definecolor{esplightpurple}{RGB}{197,137,232}
\definecolor{espdarkpurple}{RGB}{50,14,59}
\definecolor{esporange}{RGB}{255,153,0}

\begin{document}

\title{Towards computing low-makespan solutions for multi-arm multi-task planning problems}
\author{Valentin N. Hartmann$^{1, 2}$, Marc Toussaint$^{2}$\\%
$^{1}$Visus, University of Stuttgart, Germany, 
    { valentin.hartmann@ipvs.uni-stuttgart.de}\\ 
$^{2}$Learning and Intelligent Systems Lab, TU Berlin, Germany
}

\maketitle

\begin{abstract}

We propose an approach to find low-makespan solutions to multi-robot multi-task planning problems in environments where robots block each other from completing tasks simultaneously.

We introduce a formulation of the problem that allows for an approach based on greedy descent with random restarts for generation of the task assignment and task sequence.
We then use a multi-agent path planner to evaluate the makespan of a given assignment and sequence.
The planner decomposes the problem into multiple simple subproblems that only contain a single robots and a single task, and can thus be solved quickly to produce a solution for a fixed task sequence.
The solutions to the subproblems are then combined to form a valid solution to the original problem.

We showcase the approach on robotic stippling and robotic bin picking with up to 4 robot arms.
The makespan of the solutions found by our algorithm are up to 30\% lower compared to a greedy approach.
\end{abstract}
\section{Introduction}

In many common applications, a given set of tasks need to be completed.
Examples include, e.g., unloading a dishwasher, laser welding, bin picking, inspection, or unloading a delivery truck.
Such problems could often benefit from parallelization of tasks by using multiple robots.
This requires deciding which robot works on which tasks, and in which order the tasks should be completed to minimize the total needed time.
Both these things are nontrivial problems.

For a single robot, similar problems were previously solved in laser welding (\cite{saha2006planning,kovacs2016integrated}) by formulating them as traveling salesman problem with neighborhoods.
Problems involving multiple mobile robots have been tackled by treating the problem as search on graphs (\cite{yu2016optimal,ma2016optimal}) or with conflict based search (\cite{honig2018conflict}).

A common assumption in planning for multi-agent systems in task and motion planning is synchronicity of tasks, i.e., each agent starts and finishes its task at the same times (\cite{pan2021general,toumi2022multi,zhang2022mip}).
Planning with this synchronicity assumption does not work well if tasks have different lengths and leads to idle times for the robots if they block each other from starting a task.


\cite{chen2022cooperative} tackle finding low-makespan solutions for assembly scenarios, i.e. scenarios where the order of tasks is highly constrained using mixed integer linear programs to compute the task assignment.
However, the resulting computation times are in the thousands of hours, and the algorithm has only been demonstrated for up to three robots.

To tackle multi robot task assignment for offshore missions, temporal planning is used in \cite{carreno2020task}, but the robots tend not to interfere with each other much, whereas in our setting, the space is much more congested.

Finally, we refer to \cite{stern2019multi} for a survey on multi robot motion planning.
We particularly want to point out priority based search \cite{hang2019prio}, which is most similar to our approach.
However, we do not introduce the priority on the path-level, but on the tasks for each robot.


To compute a good solution to the combined multi-agent task assignment and path planning problem, we decompose the task assignment and the multi-agent planning problems.
Particularly, we serialize the multi-robot task sequence to obtain a sequence that defines precedence constraints and assigns a robot to each task.
For this sequence, we then plan a path to determine its makespan.
This enables us to proceed with a standard greedy search with random restarts to find an improved sequence.


We evaluate this approach on robotic stippling and robotic bin-picking with multiple robots and compare it to a greedy approach and an approach using only a single arm.
\section{Problem Formulation \& Notation}

We consider settings with $R$ robots, indexed by $r\in\{1, ..., R\}$, with configuration spaces $\mathcal{C}^r$, and $M$ objects, indexed by $m\in\{1, ..., M\}$ with poses in $SE(3)$.
We use $\mathcal{X} = \mathcal{C}^1\times...\times\mathcal{C}^R\times SE(3)^M$ to denote the configuration space and $x:[0, t_\text{final}]\rightarrow\mathcal{X}$ to denote a path through this configuration space.
We use $s^r = \left(s_1^r, ..., s_{N_r}^r\right)$ to denote the sequence of tasks that robot $r$ needs to fulfil, and $s=s^1\times ...\times s^R\in\mathbb{S}$ for the combined sequence containing every robot.
The set $\mathbb{S}$ consists of all valid sequences that complete all $N$ tasks.
The tasks in $s^r$ must be completed by robot $r$ in the order in which they are contained in $s^r$, i.e., $T^r_i<T^r_{i+1}$, where $T^r_i$ is the time at which robot $r$ finishes task $s_i^r$.
Similar to $s$, we use $T$ do denote all finishing times $T = T^1\times ... \times T^R$, with $T^r = (T_1^r, ..., T_{N_r}^r)$.
We want to solve
\begin{subequations}\label{eq:eq}
\begin{align}
    \min_{s, x, T}\, & \max_{r}\,T_{N_r}^r \\
    \text{s.t.}\, & x(0) = x_0\\
    \forall r\, & g(x^r(t), s_i^r) \leq 0 \,\, t\!\in\![T^r_{i-1}, T^r_i]\,\forall i\!\in\!(1, ..., N_r)\label{eq:task_constraint}\\
    \forall r\, & h(x^r(T^r_i), s_i^r) = 0 \,\, \forall i\!\in\!(1, ..., N_r)\label{eq:task_goal_constraint}\\
    & x(t) \in \mathcal{C}_\text{free}\,\, \forall t\in[0, \max_{r}\,T_{N_r}^r]\label{eq:coll_free}\\
    & s \in \mathbb{S}\label{eq:logic},
\end{align}
\end{subequations}
i.e., want to find a sequence of task-assignments and the corresponding paths that minimize the makespan.

In \eqref{eq:eq}, \eqref{eq:task_constraint} are the constraints on the path that are active in the current task (e.g., robot $r$ is moving object $m$), \eqref{eq:task_goal_constraint} constrains the path of robot $r$ such that task $s_i^r$ is completed at time $T_i^r$ via the constraint function $h$ (e.g., a pose-constraint on object $m$), \eqref{eq:coll_free} ensures that the found path is collision free, and \eqref{eq:logic} ensures that the sequence of tasks is a valid sequence that fulfills all tasks (and possibly constrains the order of tasks).
This is a Logic-Geometric program (\cite{18-toussaint-RSS}).

This problem could be formulated as a multi-agent traveling salesman problem, and could theoretically be solved using mixed integer programming.
However, the path planning problem for a multi-robot system itself is hard, and a purely optimization based approach is thus unsuitable, as many of the local optima are infeasible due to violation of \eqref{eq:coll_free}



In the following, we assume that:
\begin{itemize}
    \setlength\itemsep{0em} 
    \item Each task can be handled by one robot, and
    \item There exists a `resting' pose for each robot in which no other robot is blocked from completing its tasks, where the robot can return to after competing its task.
\end{itemize}
The first assumption is made in this work for simplicity, but can be removed by enabling planning for multiple robots for a task (as done in \cite{hartmann2021long}).
The second assumption enables completeness of the planning subroutine.

\section{Method}
To make the problem tractable, we decompose the problem into `generating a robot-task sequence' and `evaluating the sequence'. 
It is then possible to use search methods, e.g., greedy descent to improve an initial sequence.

\begin{figure}[t]
    \centering

    \includegraphics[width=.97\linewidth]{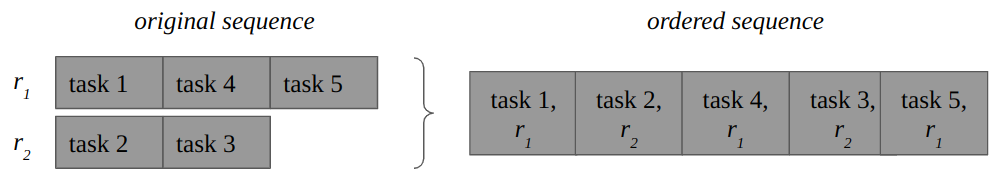}
    \caption{Example of a serialization $\hat{s}$ of a sequence $s$.}

    \label{fig:flattening}
\end{figure}

While $s$ determines the sequence of tasks per robot $r$, it does not determine their relative ordering across robots.
Instead of searching over $s$, we search over a \emph{serialized} sequence $\hat{s}$ (see \cref{fig:flattening} for an illustration).
If $s$ is partially ordered, i.e., the tasks have dependencies between each other, such a serialization can be obtained via topological search.

Given $s$, such a serialized assignment is defined as $\hat{s} = (s_{i(j)}^{r_{k(j)}}, r_{k(j)})_{j=0}^N$, where $k(j)$ maps the index $j$ of a task in sequence $\hat{s}$ to the index of the robot that the task is assigned to, and $i(j)$ maps the index $j$ of the sequence $\hat{s}$ to the index of the sequence $s^r$. 
The ordering in the serialized sequence imposes that task $s_{i(j)}^{r_{k(j)}}$ must be finished before task $s^{r_{k(j+1)}}_{i(j+1)}$, i.e., $T_{i(j)}^{r_{k(j)}} < T_{i(j+1)}^{r_{k(j+1)}}$.
For any sequence $s$, many possible serializations $\hat{s}$ exist.
As we search over $\hat{s}$, we can cover all possible precedence constraints.


We then compute a valid path for the multi-agent system for a serialized sequence using an approach similar to \cite{hartmann2021long}, i.e., by planning the tasks in $\hat{s}$ one-by-one, and treating the previously computed paths as fixed.

\begin{algorithm}[t]
\caption{Greedy descent with restarts for multi robot task planning} 
\label{alg:sim_annealing}
    $x_\text{min}\gets\emptyset \,\, t_\text{min}\gets \infty$\;
    \For{$j\gets0$ \KwTo \texttt{max\_outer\_iter}\label{alg:greedy_outer}}{
        $\hat{s}\gets \texttt{InitializeSequence}$()\label{alg:init_sequence}\;
        $x_\text{inner, min}, t_\text{inner, min}\gets \texttt{PlanGivenSequence($\hat{s}$)}$\label{alg:plan_seq}\;
        \For{$i\gets0$ \KwTo \texttt{max\_inner\_iter}}{
            $\hat{s}_\text{cand}\gets \texttt{GenerateNeighbor}$($\hat{s}$)\label{alg:neighbor}\;
            $x, t\gets \texttt{PlanGivenSequence}(\hat{s}_\text{cand})$\;
            \If{$t<t_\text{inner, min}$\label{alg:inner_update}}{
                $x_\text{inner, min}, t_\text{inner, min}\gets x, t$\;
                $\hat{s}\gets\hat{s}_\text{cand}$\;
            }
            \If{$t<t_\text{min}$\label{alg:outer_update}}{
                $x_\text{min}, t_\text{min}\gets x, t$\;
            }
        }
    }
    \Return $x_\text{min}, t_\text{min}$
\end{algorithm}


\begin{algorithm}[t]
\caption{\texttt{GenerateNeighbor}$(\hat{s})$} 
\label{alg:sequence}
    $r\gets \texttt{rnd}(0,1)$\;
    \lIf{$r < \frac{1}{3}$}{
        \Return \texttt{SwapRobot($\hat{s}$)} \label{alg:swap_robot}
    }
    \lElseIf{$\frac{1}{3} \leq r < \frac{2}{3}$}{
        \Return \texttt{SwapRandomElements($\hat{s}$)} \label{alg:swap_element}
    }
    \lElse{
        \Return \texttt{ReverseSubtour($\hat{s}$)} \label{alg:rev_subtour}
    }
\end{algorithm}

\begin{algorithm}[t]
\caption{$\texttt{PlanGivenSequence}(\hat{s})$} 
\label{alg:planning}
    $\texttt{sol}\gets\{\}$\;
    \For{$i\gets0$ \KwTo len($\hat{s}$)}{
        $(x_i, t_i)\gets$\texttt{PlanTaskForRobot}($\hat{s}_i, \texttt{sol}$)\;
        $\texttt{sol}\gets (x_i, t_i)$\label{alg:store_sol}\;
    }
    $(x, T) \gets$ \texttt{GetPathsAndMakespan}($\texttt{sol}$)\;
    \Return $(x, T)$
\end{algorithm}




\subsection{Time embedded multirobot path planning}
To compute a plan for a given sequence, we build on the work from \cite{hartmann2021long}:
In the path planning subroutine, we decompose the full problem, and solve sub-problems with a single robot and a single task only.
That is, we sequentially plan the robot/task-pairings in $\hat{s}$, while treating previously planned paths as fixed.
The precedence constraints encoded in $\hat{s}$ are respected in this planning step, i.e., we enforce that the tasks are fulfilled in the order specified in $\hat{s}$ by introducing a lower bound on the finishing time of the plan, i.e., $T^r_{i} < T^r_{i+1}$.

\begin{figure}[t]
\centering

    \includegraphics[width=.97\linewidth]{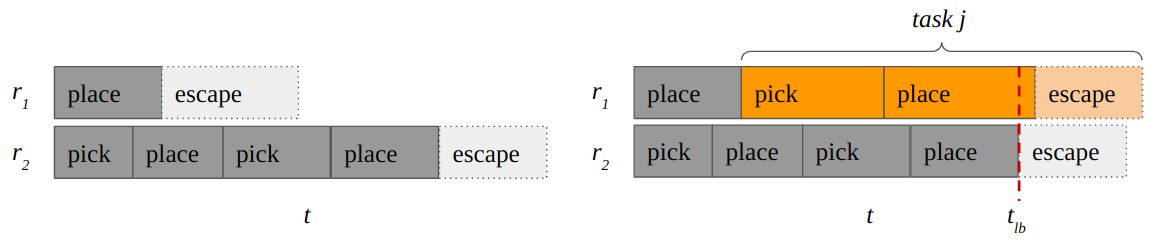}
    \caption{Illustration of a single planning step in the sequential planning for planning task $j$: a pick and place task consisting of a `pick' and a `place' action.
    }
    \label{fig:planing_illustration}
\end{figure}

After completion of a task, an `escape' path is planned to the pose that does no block any other robot.
Being able to move to this non-blocking pose guarantees that there is always a valid path for subsequent tasks and robots, since there is a time in the future where all robots that are not involved in the current task are at their non-blocking pose.
This escape path is discarded when planning the paths for the next task for the robot.
\Cref{fig:planing_illustration} illustrates how the paths are planned using time-embedded path planning.

We use ST-RRT* (\cite{22-grothe-ICRA}) which allows the specification of a maximum velocity for time-embedded path planning, and run it until a solution was found, and continue optimizing the path for a fixed number of iterations.
This results in a good path, but typically not in the optimal one due to the restriction of the runtime.
We post-process the resulting path via shortcutting (\cite{hauser2010fast} adapted to the time-embedded case), and smooth it with an optimizer to minimize accelerations.
This results in paths that can be directly executed on a real robot.

\subsection{Algorithms \& Practical details}

We give pseudocode for the proposed approach in  \cref{alg:sim_annealing,alg:sequence,alg:planning}.
We implement a greedy descent with restarts, i.e., the greedy descent in the inner loop is repeated \texttt{max\_outer\_iter} times (\cshref{alg:sim_annealing}, \cref{alg:greedy_outer}).
The serialized assignment is randomly initialized (\cshref{alg:sim_annealing}, \cref{alg:init_sequence}) before planning for the given sequence (\cshref{alg:sim_annealing}, \cref{alg:plan_seq}).
In the inner loop, a neighbouring sequence is generated (\cshref{alg:sim_annealing}, \cref{alg:neighbor}), and a plan is generated.
If the makespan is lower than the current best makespan that was found in the inner loop so far, the best makespan of the inner loop, the corresponding path, and the current best serialized sequence are updated (\cshref{alg:sim_annealing}, \cref{alg:inner_update}).
If the makespan is lower than the best makespan achieved so far, the best overall makespan and corresponding path are updated.
Planning the path for a given sequence $\hat{s}$ proceeds iteratively for each task (\cshref{alg:sim_annealing}, \cref{alg:outer_update}) until all tasks are planned.
Previous solutions are respected in the planning process by storing the previous solutions (\cshref{alg:sim_annealing}, \cref{alg:store_sol}), and considering them as constraints for subsequent tasks.

We make the following implementation choices, and small modification to the core idea presented before to speed up the search and planning process:

\paragraph{Generating the initial sequence}\label{ssec:initial_sec}
We generate the initial sequence $\hat{s}$ by going through the robots in round-robin, and assigning a random feasible task to the current robot.
This results in all the agents alternating between each other.

\paragraph{Altering a sequence}
Greedy descent requires a function that generates a `neighbouring' sequence from a given sequence\footnote{If precedence constraints are present in the original sequence $s$ (e.g. when stacking boxes, the lowest box must be placed first), these constraints could be enforced by, e.g., rejection sampling.}.
We do this with equal probability by either (i) Swapping elements of the sequence (\cshref{alg:sequence}, \cref{alg:swap_element}), (ii) Reversing subtours (\cshref{alg:sequence}, \cref{alg:rev_subtour}), or (iii) Swapping robots (\cshref{alg:sequence}, \cref{alg:swap_robot}).
These operations are inspired by typical approaches used in greedy searches on the traveling salesman problem (\cite{abdoun2012analyzing}).


\paragraph{Early stopping of motion planning}
A lower bound on the makespan of the remaining sequence can be computed by using the euclidean distance between poses and optimistically assume movement at maximum velocity to obtain a lower bound on the finishing time of a task.
This lower bound can then be used to avoid the costly computation of paths that can not improve the current solution.

\paragraph{Caching of subsequences}
If we already computed a valid plan for a sequence $\hat{s}_1$, and are now planning for a sequence $\hat{s}_2$, and the first $m$ elements of two sequences $\hat{s}_1$ and $\hat{s}_2$ are the same, we can reuse the first $m$ plans that were computed for this subsequence, and only need to compute the plans for the remaining tasks.
\begin{figure}[t]
\centering
    \begin{subfigure}[t]{.13\textwidth}
        \centering
        \includegraphics[width=.97\linewidth]{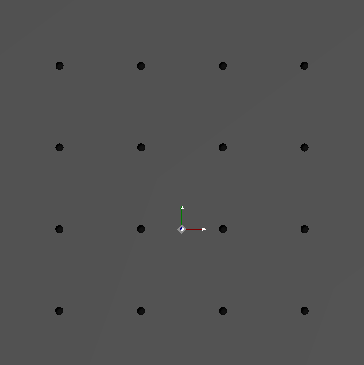}
        \caption{Grid}
    \end{subfigure}%
    \begin{subfigure}[t]{.13\textwidth}
        \centering
        \includegraphics[width=.97\linewidth]{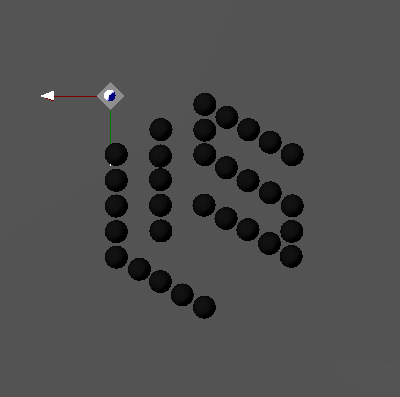}
        \caption{LIS-logo}
    \end{subfigure}%
    \begin{subfigure}[t]{.21\textwidth}
        \centering
        \includegraphics[width=.97\linewidth]{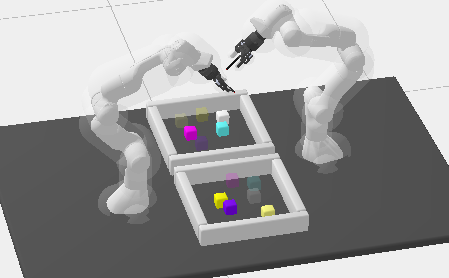}
        \caption{Bin picking}
    \end{subfigure}
    \caption{\textbf{Left} and \textbf{middle}: Positions of the tip of the pen for the stippling scenarios. \textbf{Right}: Bin picking scenario with the goal positions of the objects in low opacity.}
    \label{fig:problems}
\end{figure}

\section{Experiments}
We demonstrate our approach on the example of robotic stippling (where a task corresponds to `go to pose')\footnote{`Robotic stippling' uses robot arms equipped with pens to make dots on paper.}, and multi-arm bin picking (where one task is `pick obj, place obj', i.e., two actions).
The stippling and the bin-picking setting are shown in \cref{fig:problems}.

We analyze several versions of the stippling scenarios: for the grid, we compare a two and a four robot version; for the LIS logo, we consider a `small' version, where dots can not be made with two arms at the same time, and that can not be well parallelized, and a `large' version, where some dots can be made simultaneously.

We compare our algorithm to a baseline of a single robot, and an initialization that alternates the robot-tasks in a greedy manner, i.e., always chooses the task next that has the minimum estimated distance to the current robot pose.

Videos of the execution of some generated paths in simulation and on real robots can be found at \url{https://vhartmann.com/seq-opt}.

\begin{figure}[t]
\centering
    \begin{subfigure}[t]{.23\textwidth}
        \centering
        \includegraphics[width=.97\linewidth]{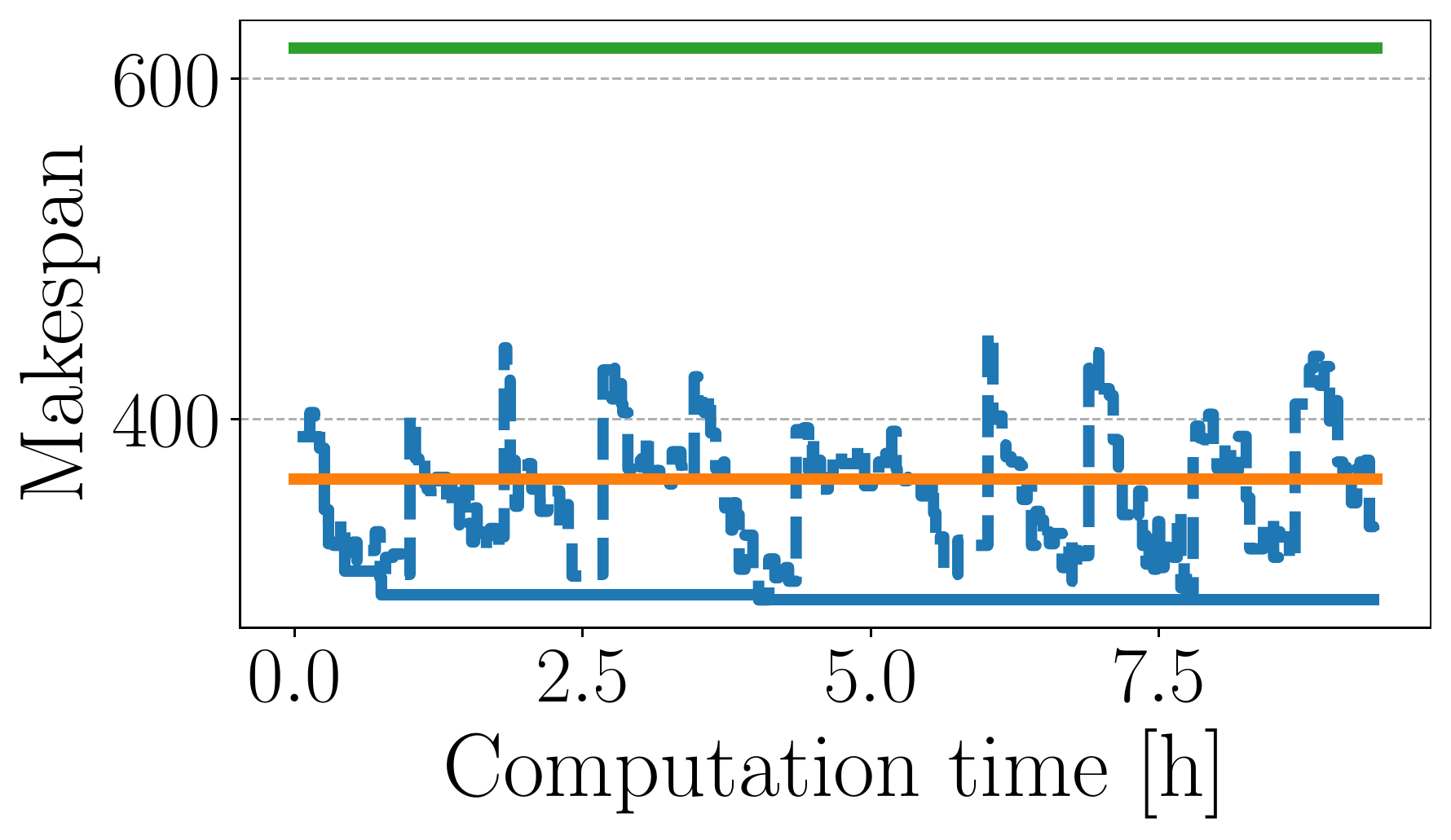}
        \caption{Grid, 2 robots}
    \end{subfigure}%
    \begin{subfigure}[t]{.23\textwidth}
        \centering
        \includegraphics[width=.97\linewidth]{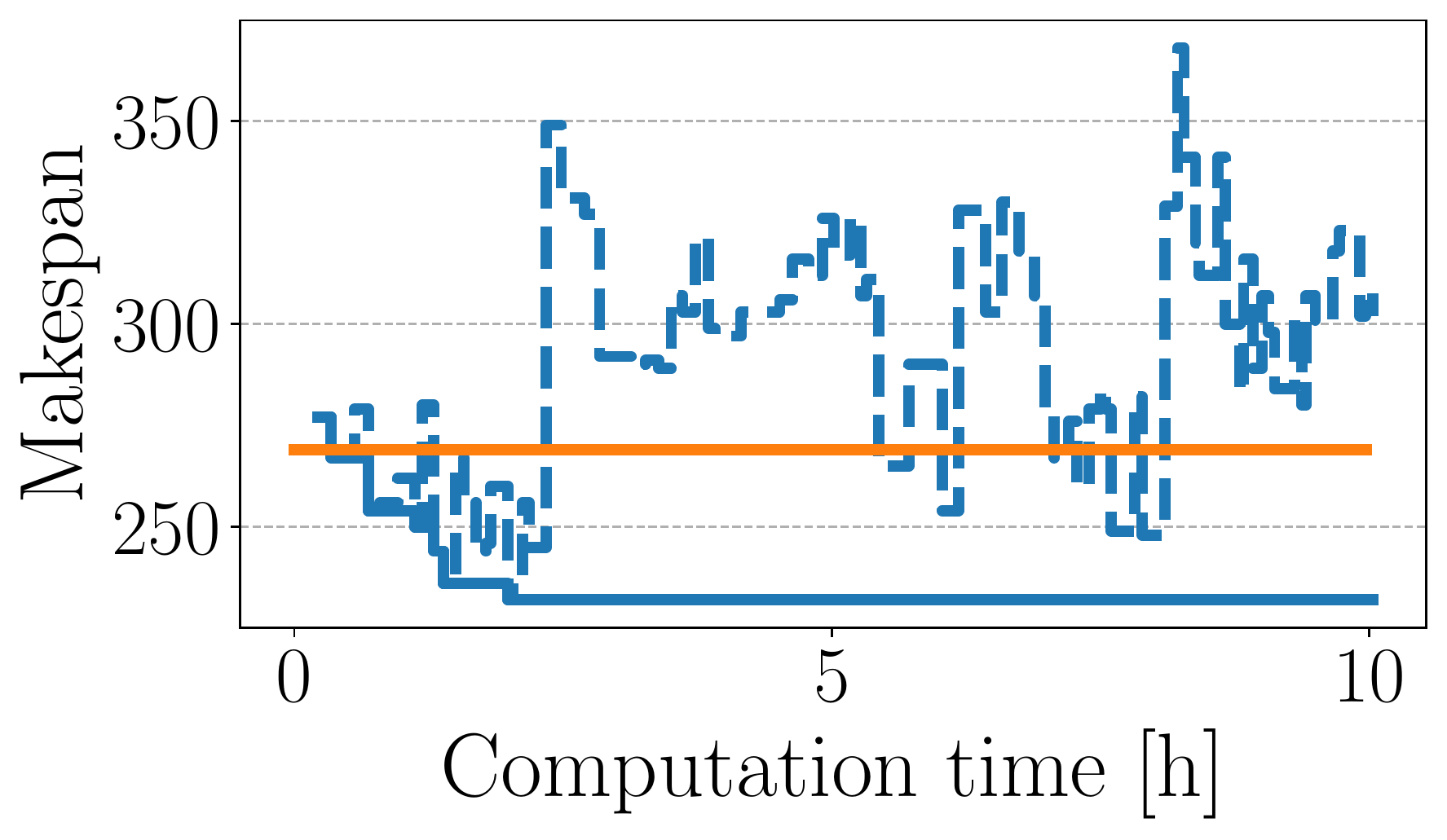}
        \caption{Grid, 4 robots}
    \end{subfigure}
    
    \begin{subfigure}[t]{.23\textwidth}
        \centering
        \includegraphics[width=.97\linewidth]{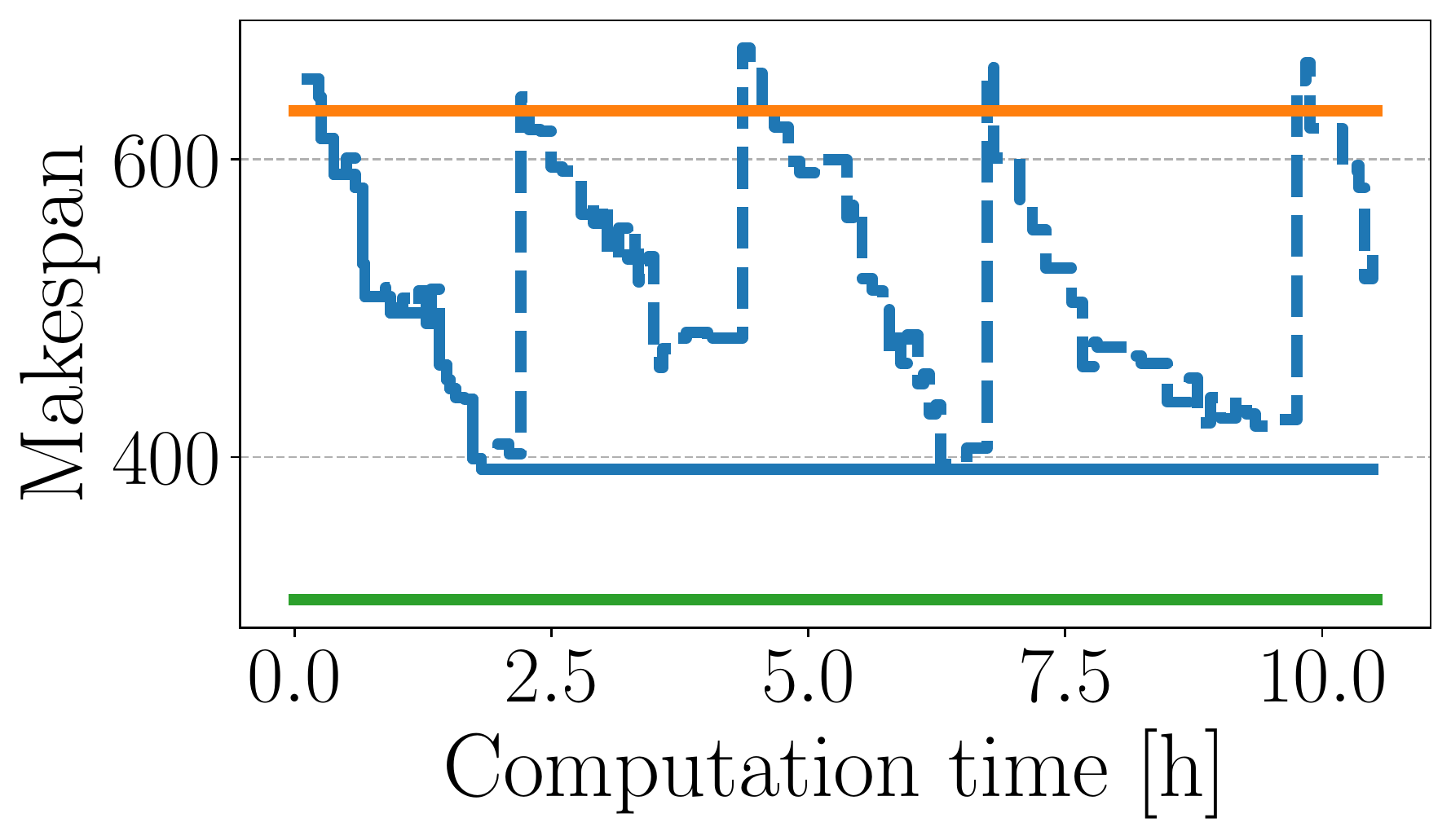}
        \caption{LIS-Logo (small)}
    \end{subfigure}%
    \begin{subfigure}[t]{.23\textwidth}
        \centering
        \includegraphics[width=.97\linewidth]{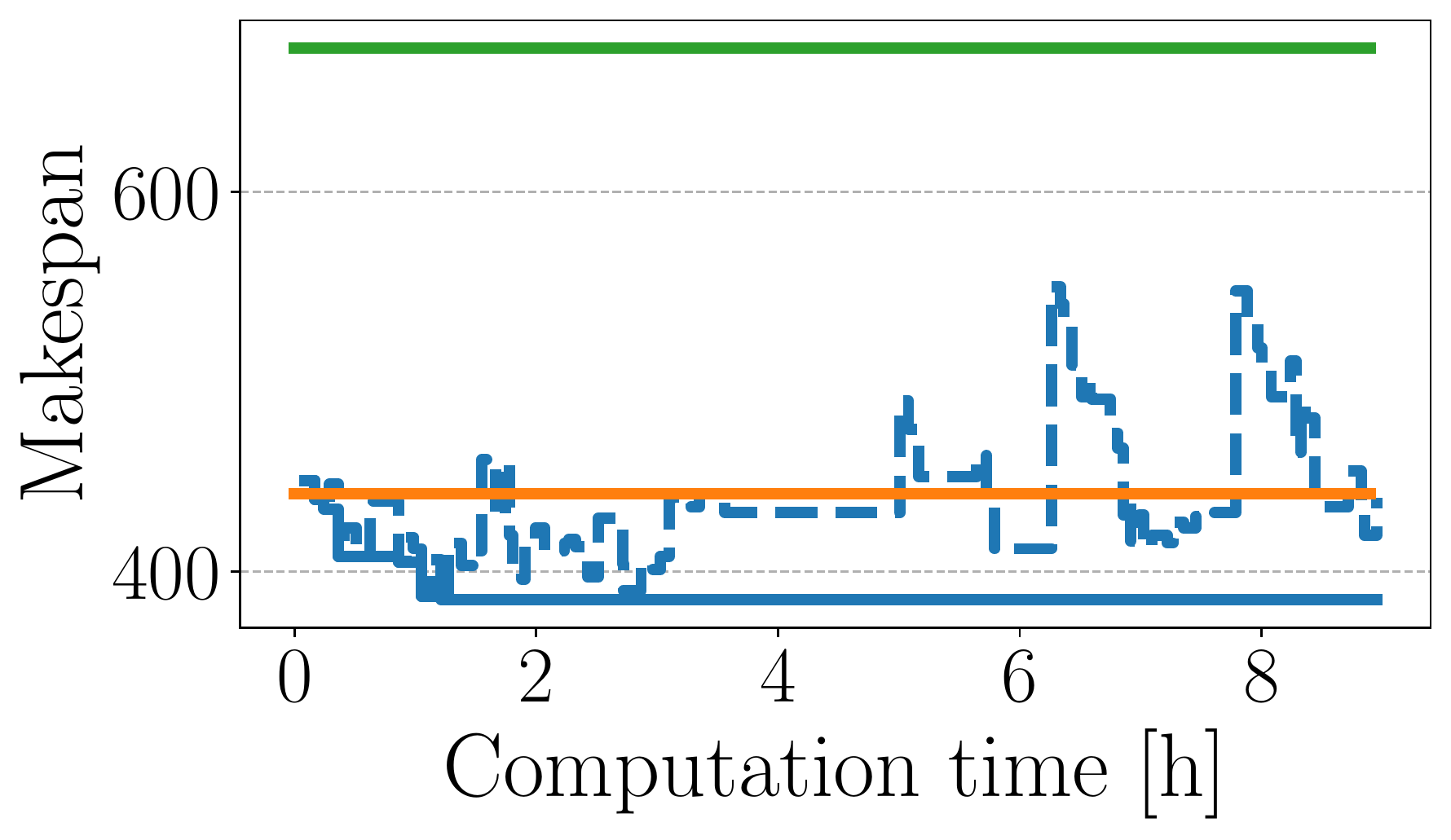}
        \caption{LIS-Logo (large)}
    \end{subfigure}
    \\[0.5em]
    \begin{subfigure}[b]{1.0\linewidth}%
        \centering
        \begin{tikzpicture}
\begin{axis} [
  width=\textwidth,
  height=0.5\textwidth,
  unbounded coords=jump,
  xtick align=inside,
  ytick align=inside,
  anchor=north,
  hide axis,
  xmajorgrids,
  ymajorgrids,
  major grid style={densely dotted, black!20},
  xmin=0,
  xmax=10,
  ymin=0,
  ymax=10,
  xlabel style={font=\footnotesize},
  xticklabel style={font=\footnotesize},
  ylabel style={font=\footnotesize},
  yticklabel style={font=\footnotesize},
  legend style={anchor=south, legend cell align=left, legend columns=3, at={(axis cs:5, 6)}, font=\small}
]
\addlegendimage{espblue, line width = 1.0pt, mark size=1.0pt, mark=square*}
\addlegendentry{Greedy Descent}
\addlegendimage{esporange, line width = 1.0pt, mark size=1.0pt, mark=square*}
\addlegendentry{Greedy}
\addlegendimage{espgreen, line width = 1.0pt, mark size=1.0pt, mark=square*}
\addlegendentry{Single Arm}

\end{axis}
\end{tikzpicture}
    \end{subfigure}
    \vspace{-5mm}
    \caption{
    Makespan of current sequence (dashed line), and minimum makespan so far (solid line) at a computation time on the stippling scenarios.
    Both greedy, and single arm are not anytime, so do not improve the initial solution.
    For the four arm grid setting, there is no comparisons to the single arm baseline, as not all dots can be made by all robots.
    }
    \label{fig:cost_plot_stippling}
\end{figure}

\subsection{Results}
\Cref{fig:cost_plot_stippling} shows the evolution of the makespan over the computation time for the stippling scenarios.
For the four-arm scenario, the single-arm baseline is not available, as not every dot can be made by every arm, and the baseline-sequence is thus not feasible.

\begin{figure*}[t]
\centering
    \begin{subfigure}[t]{.19\textwidth}
        \centering
        \includegraphics[width=.97\linewidth]{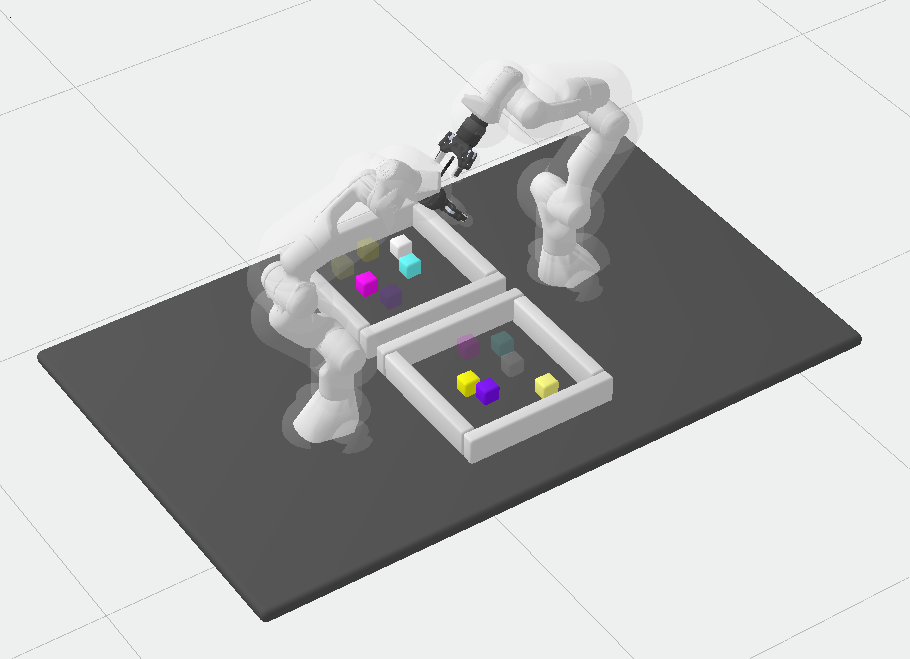}
    \end{subfigure}%
    \begin{subfigure}[t]{.19\textwidth}
        \centering
        \includegraphics[width=.97\linewidth]{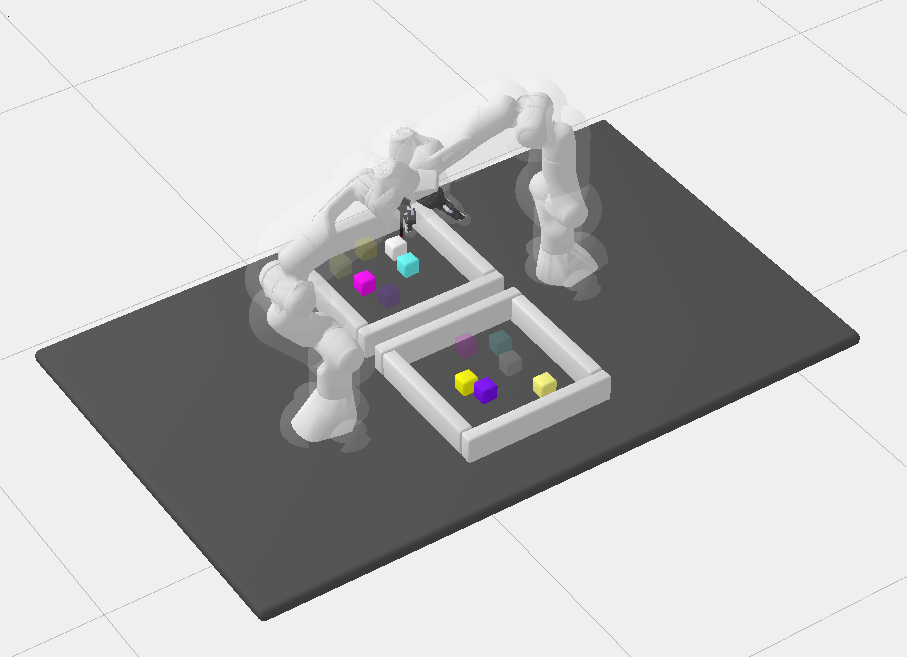}
    \end{subfigure}%
    \begin{subfigure}[t]{.19\textwidth}
        \centering
        \includegraphics[width=.97\linewidth]{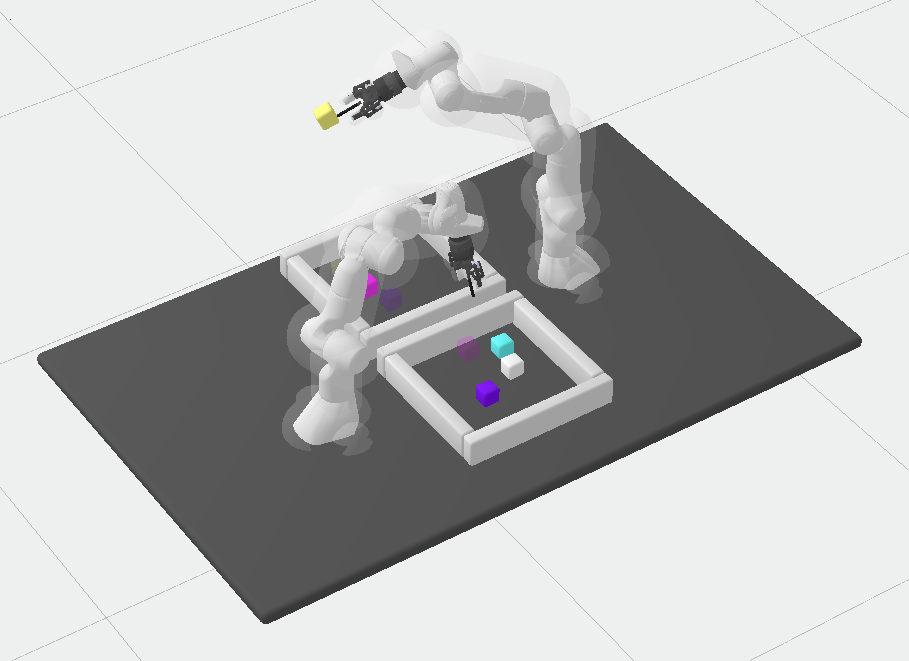}
    \end{subfigure}%
    \begin{subfigure}[t]{.19\textwidth}
        \centering
        \includegraphics[width=.97\linewidth]{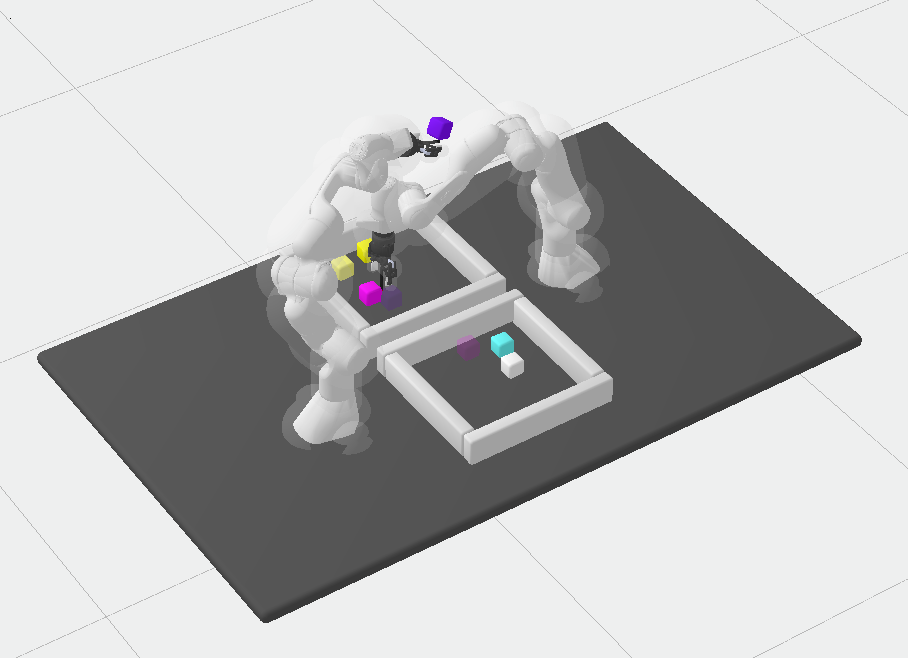}
    \end{subfigure}%
    \begin{subfigure}[t]{.19\textwidth}
        \centering
        \includegraphics[width=.97\linewidth]{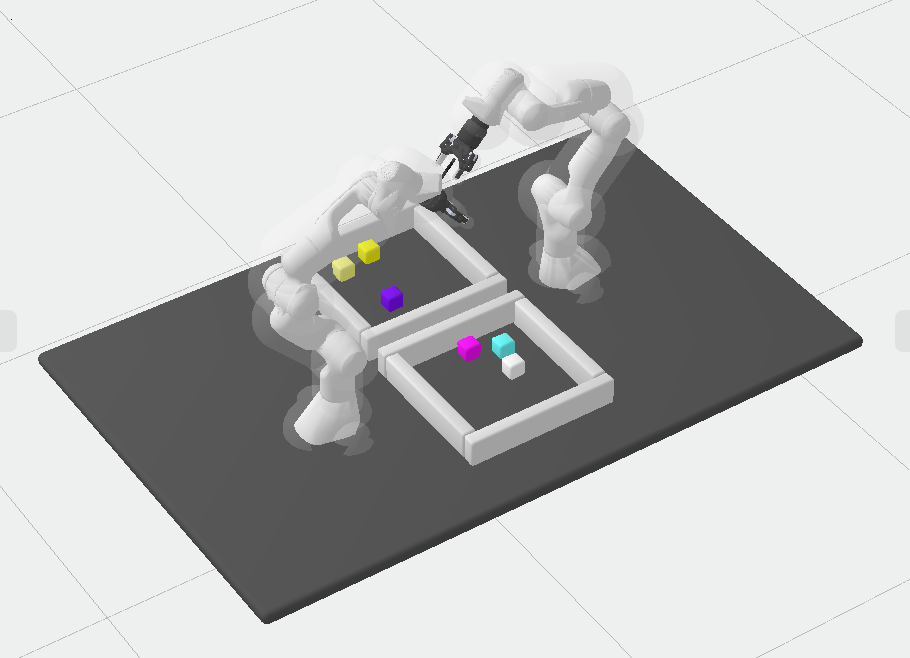}
    \end{subfigure}
    \caption{Snapshots of two robots working through a task sequence from the bin picking scenario.}
    \label{fig:sequence}
\end{figure*}

 \begin{figure}[t]
\centering
    \includegraphics[width=.3\textwidth]{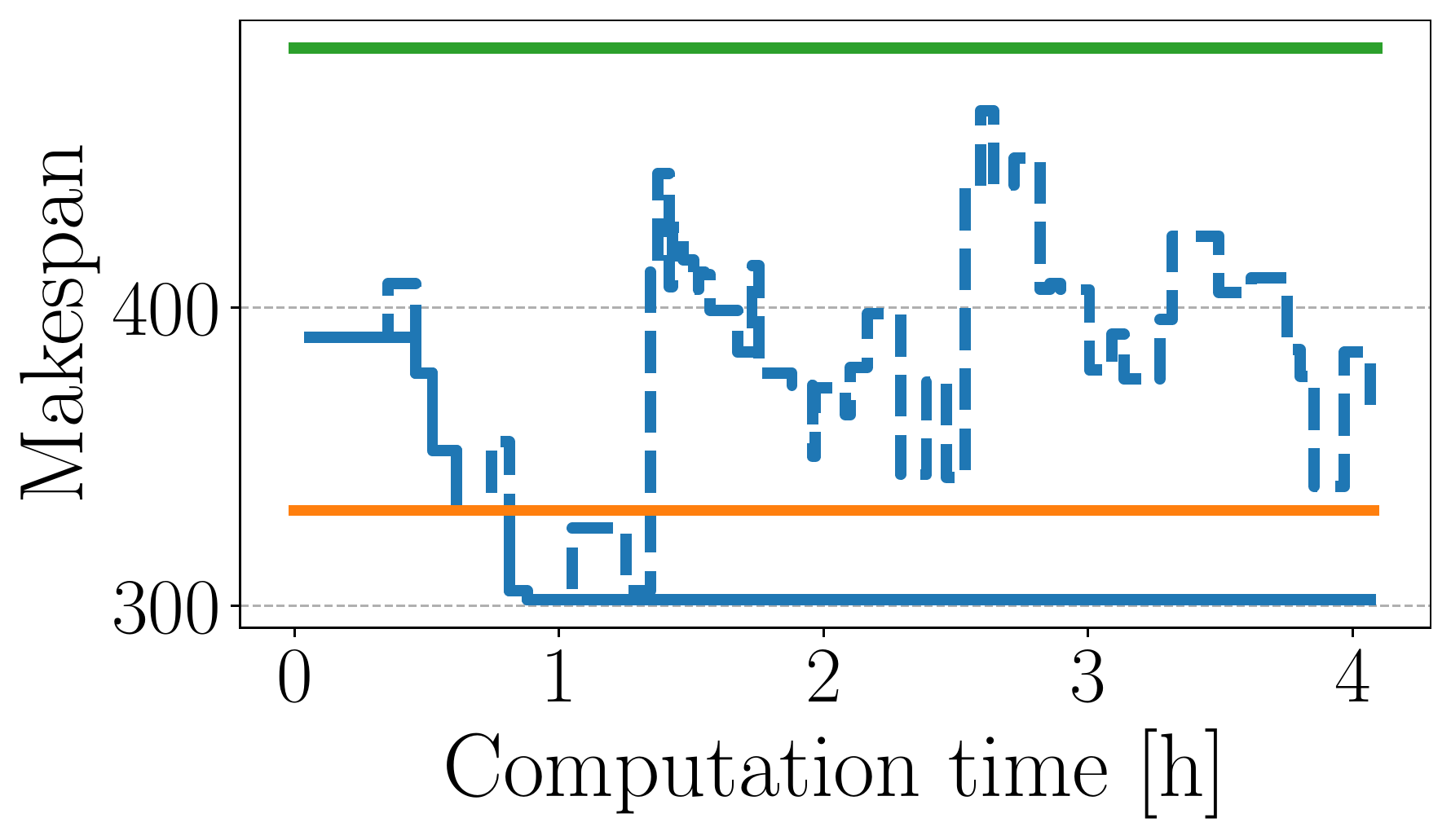}
    \begin{subfigure}[b]{1.0\linewidth}%
        \centering
        \begin{tikzpicture}
\begin{axis} [
  width=\textwidth,
  height=0.5\textwidth,
  unbounded coords=jump,
  xtick align=inside,
  ytick align=inside,
  anchor=north,
  hide axis,
  xmajorgrids,
  ymajorgrids,
  major grid style={densely dotted, black!20},
  xmin=0,
  xmax=10,
  ymin=0,
  ymax=10,
  xlabel style={font=\footnotesize},
  xticklabel style={font=\footnotesize},
  ylabel style={font=\footnotesize},
  yticklabel style={font=\footnotesize},
  legend style={anchor=south, legend cell align=left, legend columns=3, at={(axis cs:5, 6)}, font=\small}
]
\addlegendimage{espblue, line width = 1.0pt, mark size=1.0pt, mark=square*}
\addlegendentry{Greedy Descent}
\addlegendimage{esporange, line width = 1.0pt, mark size=1.0pt, mark=square*}
\addlegendentry{Greedy}
\addlegendimage{espgreen, line width = 1.0pt, mark size=1.0pt, mark=square*}
\addlegendentry{Single Arm}

\end{axis}
\end{tikzpicture}
    \end{subfigure}
    \vspace{-5mm}
    \caption{
    Makespan of current sequence (dashed line), and minimum makespan so far (solid line) at a computation time on the bin-picking scenario.
    }
    \label{fig:cost_plot_bin_picking}
\end{figure}

The average planning time for an individual sequence was roughly 180s for the grid with two robots (11s per task), 543s for the grid with four robots (33s per task), 288s for the small logo (9.6s per task) and 185s for the large logo (6s per task).

\Cref{fig:sequence} shows a sequence of snapshots from two robots executing an optimized sequence from the bin-picking scenario, and \cref{fig:cost_plot_bin_picking} shows the evolution of the makespan.
Here, the average planning time for an individual sequence was roughly 110s 
 (consisting of 6 tasks, each containing 2 actions), i.e. 18s per task, respectively 9s per action.

In these experiments, it can be seen that the improvement of using multiple arms is more pronounced in the case where tasks can be completed in parallel (i.e. in the grid, the large LIS logo, and the bin-picking scenarios) compared to the case where the arms completely block each other (small LIS logo), where the single-arm version is quicker than the one using two arms.
In this case, our algorithm finds a solution that minimizes the `switches' between robots as well, but does not completely reach it due to the limited number of iterations of the inner loop of the search.

Comparing the results of the grid with two and four arms, it can be seen that the planning time increases considerably, as the space becomes more occupied by the robots, and the planning queries become more complex (and thus slower).
The algorithm is still able to handle this, and finds a good solution, even though many fewer iterations are done in the same computation time.

\section{Discussion \& Limitations}
Both multi-agent task assignment and planning are hard problems.
We showed the algorithm on multiple scenarios with up to four robots.
The setting with four robots is extremely crowded, and planning times are thus high, which leads to less iterations of the search algorithm.
The approach we show here works well in practice for our problems, even though it is not time-optimal.

To achieve time-optimal plans, one would have to take into account dynamics, and consider an approach that enables joint planning of multiple agents.
Settings where many robots are trying to occupy the same space (as the four arm stippling-scenario) could benefit most from such a joint approach.

We noted that the lower bound that we compute tends to be optimistic, and does not help much in reducing computation time.
It might be worth the effort to compute a more accurate lower bound for the time it takes to complete a task.

We currently only consider single poses to fulfill a task, and do not allow `neighborhoods', i.e., the set of poses that fulfill the constraints associated with a task.
We expect this to be interesting for future work.

Further, we want to investigate more informed ways of generating candidate sequences that take into account the geometry of the scene, and the tasks that should be done by each robot.

Finally, we restricted our problem setting here to tasks that can be solved by a single robot.
In the future, we want to investigate more complex task and motion planning settings, e.g. tasks that require cooperation, or non-prehensile manipulation.





\section{Conclusion}
We presented an approach that enables the optimization of the task sequence and assignment in multi-agent systems.
The approach can lead to improvements of the makespan of up to 50\% over a single robot doing all tasks, and up to a 30\% improvement over a greedy alternating assignment.

We believe that the work can benefit from more research on how to speed up the multi-agent path planning, and from exploring other search approaches that account for the computationally expensive motion planning.


\section*{Acknowledgements}
\footnotesize{
The authors would like to thank Wolfgang Hönig for helpful comments on the draft of this work.
We would also like to thank the anonymous reviewers for their helpful comments.

The research has been supported by the Deutsche Forschungsgemeinschaft (DFG, German Research Foundation) under Germany's Excellence Strategy -- EXC 2120/1 -- 390831618, and the DAAD.
}


\bibliography{ref} 
\bibliographystyle{aaai}

\end{document}